\documentclass{article}

\usepackage{arxiv}

\usepackage[utf8]{inputenc} 
\usepackage[T1]{fontenc}    
\usepackage{hyperref}       
\usepackage{url}            
\usepackage{booktabs}       
\usepackage{amsfonts}       
\usepackage{nicefrac}       
\usepackage{microtype}      
\usepackage{lipsum}
\usepackage{graphicx}
\graphicspath{ {./images/} }

\usepackage{graphicx}%
\usepackage{multirow}%
\usepackage{amsmath,amssymb,amsfonts}%
\usepackage{amsthm}%
\usepackage{mathrsfs}%
\usepackage[title]{appendix}%
\usepackage{xcolor}%
\usepackage{textcomp}%
\usepackage{manyfoot}%
\usepackage{booktabs}%
\usepackage{algorithm}%
\usepackage{algorithmicx}%
\usepackage{algpseudocode}%
\usepackage{listings}%
\usepackage{array}
\usepackage{siunitx}
\definecolor{newcolor}{rgb}{.8,.349,.1}
\usepackage{etoolbox}           
\renewcommand{\bfseries}{\fontseries{b}\selectfont} 
\robustify\bfseries             
\newrobustcmd{\B}{\bfseries}    

\def\BibTeX{{\rm B\kern-.05em{\sc i\kern-.025em b}\kern-.08em
    T\kern-.1667em\lower.7ex\hbox{E}\kern-.125emX}}

\title{Robust Isolation Forest using Soft Sparse Random Projection and Valley Emphasis Method}

\author{ 
 Hun Kang \\  
  Department of Data Science \\  
  Seoul National University of Science \& Technology (SeoulTech) \\  
  232 Gongreungno, Seoul 01811, Republic of Korea \\  
  \texttt{hunkang10@ds.seoultech.ac.kr} \\  
   \And  
  Kyoungok Kim \thanks{Corresponding author.} \\  
  Department of Industrial Engineering \\  
  Seoul National University of Science \& Technology (SeoulTech) \\  
  232 Gongreungno, Seoul 01811, Republic of Korea \\  
  \texttt{kyoungok.kim@seoultech.ac.kr}  
}

\begin{document}
\maketitle
\begin{abstract}
Isolation Forest (iForest) is an unsupervised anomaly detection algorithm designed to effectively detect anomalies under the assumption that anomalies are ``few and different." Various studies have aimed to enhance iForest, but the resulting algorithms often exhibited significant performance disparities across datasets. Additionally, the challenge of isolating rare and widely distributed anomalies persisted in research focused on improving splits. To address these challenges, we introduce Robust iForest (RiForest). RiForest leverages both existing features and random hyperplanes obtained through soft sparse random projection to identify superior split features for anomaly detection, independent of datasets. It utilizes the underutilized valley emphasis method for optimal split point determination and incorporates sparsity randomization in soft sparse random projection for enhanced anomaly detection robustness. Across 24 benchmark datasets, experiments demonstrate RiForest's consistent outperformance of existing algorithms in anomaly detection, emphasizing stability and robustness to noise variables.
\end{abstract}

\renewcommand{\thefootnote}{} 
\footnotetext{Preliminary work. Under review.}

\section{Introduction}
\label{sec:intro}

An anomaly refers to distinctive patterns within data that deviate from the normative behavior \cite{Chandola2009anomaly}. Such anomalies may stem from previously unrecognized processes and hold potential for transformation into valuable, actionable insights \cite{Chalapathy2019deep}. Consequently, various sectors have committed substantial efforts not only to detect anomalies but also to respond effectively to their occurrence. This commitment is evident in applications like credit card fraud detection \cite{Tran2018real, kalid2020a}, network intrusion detection \cite{aryal2021usfad, nalini2024enhancing}, hyperspectral anomaly detection \cite{Li2020hyperspectral,yuan2019hyperspectral}, manufacturing defect identification \cite{Puggini2018an,sarda2021a}, and video surveillance \cite{San2019outlier, he2024video}. The integration of the Internet of Things, cloud computing, and smart devices has driven a rapid increase in data generation, further intensifying the need for effective anomaly detection. This evolving landscape has heightened interest in anomaly detection across a wider range of domains \cite{Lazarevic2005feature,Souiden2022a}.

Among various approaches, unsupervised anomaly detection algorithms leverage the inherent characteristics of data without relying on labels. In real-world applications, acquiring labeled datasets is challenging, and generating labels for anomalies can be costly \cite{Fiore2013network}. Additionally, the considerable imbalance between anomalies and normal instances complicates the application of supervised learning within imbalanced datasets. Consequently, the development of anomaly detection algorithms has primarily centered around unsupervised learning techniques.

The Isolation Forest (iForest) is an unsupervised anomaly detection algorithm designed to leverage the defining characteristic of anomalies as ``few and different.'' Initially proposed by Liu \textit{et al.} \cite{Liu2008isolation}, iForest uses binary trees to create iterative splits based on randomly selected variables and branching points. This approach assumes that data points in low-density regions are more likely to isolate closer to the root node, forming the conceptual basis of the iForest. The algorithm has gained attention for its high performance, along with lower computational complexity and memory usage compared to other density-based methods such as LOF \cite{Breunig2000lof} and COF \cite{Tang2002enhancing,Liu2008isolation,Sadaf2020intrusion,Domingues2018a}. Additionally, iForest demonstrates rapid convergence in anomaly detection even with smaller parameter values for the number of trees and subsample size, making it especially useful in real-world settings where labeled data is limited or hyperparameter tuning is challenging \cite{Liu2008isolation,Tao2018a}.

Despite these advantages, iForest has certain limitations that have driven research efforts aimed at improving its performance.
Numerous studies focus on refining the inherent randomness within iForest and identifying optimal features (hyperplanes) and split points for more effective anomaly detection \cite{Liu2018an, Liu2010on, Liao2019entropy, karczmarek2020k, karczmarek2021fuzzy, tokovarov2022a}. These methods often suggest criteria for selecting features and split points that better separate anomalies from normal samples, in contrast to the uniform random selection of splitting points at each internal node. For instance, Liu \textit{et al.} \cite{Liu2018an} introduced the Optimized iForest (OiForest), which selects variables and split points by evaluating factors like the mean distance between left and right regions, the sum of variances in these regions, and the variance within variables. Similarly, Liu \textit{et al.} \cite{Liu2010on} proposed SCiForest, which leverages the average standard deviation of the split regions to guide optimal random hyperplane and split-point selection, effectively identifying densely clustered anomalies. Furthermore, Entropy iForest (EiForest) \cite{Liao2019entropy} uses a ``dimension entropy'' metric to detect variables with non-uniform distributions, branching at the midpoint of the largest empty region within the selected variable. 

Attempts to enhance anomaly isolation in the iForest have included leveraging projected spaces to address artifacts in the iForest score map function and increase the anomaly detection performance \cite{Hariri2021extended, Lesouple2021generalized, Tan2022sparse, xu2023deep, liu2024layered}. As highlighted in \cite{Hariri2021extended}, artifacts in iForest arise due to splitting on the original variables, which creates splits parallel to the axes. To mitigate this, Hariri \textit{et al.} \cite{Hariri2021extended} proposed the Extended iForest (EXiForest), which uses random hyperplanes for splitting, partially reducing artifacts in anomaly scores.  
Beyond artifact reduction, EXiForest more effectively isolates anomalies within complex distributions of normal samples by generating branches based on hyperplanes obtained through random projections. However, EXiForest can generate empty branches. Addressing this limitation, Lesouple \textit{et al.} \cite{Lesouple2021generalized} introduced the Generalized iForest (GiForest), which constructs random hyperplanes without intercepts. Subsequently, the Soft Sparse Projection iForest (SSPiForest) was developed to minimize irrelevant variables through sparse random projection \cite{Tan2022sparse}. Recently, the Deep iForest (DiForest) has been introduced, utilizing initialized deep neural networks to project data and enhance branching efficiency \cite{xu2023deep}.

Despite these advancements, existing iForest-based methods still face two primary limitations. First, their splitting criteria often fail to determine the optimal split points to distinguish effectively between the distributions of anomalies and normal samples. While some methods modify their splitting criteria to optimize split points, they either overlook data distribution or struggle with distributions that differ significantly in sample sizes, such as between anomalies and normal samples. Second, these methods are restricted in finding optimal split conditions because they use only one of either the original variables or hyperplanes generated by random projections to create branches. Approaches that rely exclusively on projected features may miss essential original attributes for anomaly separation, and vice versa. Due to these limitations, current methods exhibit reduced overall performance across diverse datasets, with some experiencing substantial performance degradation.

To address these limitations of existing methods and ensure consistently high performance across diverse datasets, this study introduces a novel iForest algorithm called the ``Robust Isolation Forest (RiForest).'' First, RiForest employs the valley emphasis method \cite{Ng2006automatic}, commonly used in image segmentation, to identify optimal split points on the selected. This method excels at identifying split points even in scenarios where the valley between distinct distributions is not clearly defined, such as unimodal distributions or those similar to unimodal shapes. Consequently, the valley emphasis method allows RiForest to reliably detect anomalies across a broader range of datasets. Second, RiForest strategically selects split features by considering both original attributes and randomly projected hyperplanes, rather than relying exclusively on one or the other. This approach not only enables effective separation of anomalies from normal samples but also ensures adaptability and robustness across varied dataset profiles, given that the optimal choice between random projection-based and attribute-based iForest methods can differ depending on the dataset.

The rest of paper is organized as follows: In Section~\ref{sec:background}, the principles of iForest and the process of anomaly detection are briefly explained. In Section~\ref{sec:method}, the proposed algorithm, RiForest, is introduced. Section~\ref{sec:experiment} presents the data and experimental methodology, while Section~\ref{sec:result} displays and discusses the experimental results. Finally, Section~\ref{sec:conclusion} provides the conclusion.

\section{Background}\label{sec:background}
This section serves as an introduction to the iForest algorithm, which has gained prominence as a crucial technique for unsupervised anomaly detection across diverse domains. 

The iForest algorithm is characterized by two distinct phases in the context of anomaly detection. In the training phase, the algorithm constructs binary trees to isolate anomalies. For this purpose, consider a training dataset $X=[x_1,\ldots,x_n]_{n \times d}^T$ consisting of $n$ data points, each with $d$ features. Every data sample is denoted as $x_i=[x_{i,1},\ldots,x_{i,d}]$. From this dataset, each binary tree is generated using a subsample $X_s$ of size $\psi$, explicitly $X_s=[x_1,\ldots,x_{\psi}]_{\psi \times d}^T$. During the construction of each tree, a random split variable $p$ is chosen from the $d$ features, and a split point $q$ is drawn from a uniform distribution within the range determined by the minimum and maximum values of the selected feature at each internal node. Once the split variable and point are established, the dataset is divided into two child nodes: data points that satisfy the condition $\{x_i| x_{i,p}\leq q\}$ are assigned to the left child node, while those satisfying $\{x_i| x_{i,p}> q\}$ are assigned to the right child node, thereby creating distinct subtrees. This recursive process continues until the data points within a node become isolated or a predetermined depth limit $l$ is reached. In iForest, the depth limit is typically defined based on the size of the subsample, such as $l=\lceil \log_2 \psi \rceil$. This parameter ensures controlled growth of the binary trees and ultimately enhances the algorithm's effectiveness in anomaly detection.

After constructing $T$ isolation trees (iTrees), iForest proceeds to the computation of anomaly scores, marking the beginning of the test phase. The anomaly score is determined through the calculation of the path length, denoted as $h(x)$, which is derived as a data point $x$ traverses each of the individual trees. The average path length $E(h(x))$, the average value of $h(x)$ from a collection of iTrees, is normalized by the theoretical average path length of an unsuccessful search within a binary search tree, denoted as $c(\psi)$, defined as follows:
\begin{equation}
    c(\psi) = 2H(\psi-1) - \frac{2(\psi-1)}{\psi}
\end{equation}
where $H(i)$ is the harmonic number, which can be estimated by $H(i)=\log i +\gamma$, and $\gamma$ represents the Euler constant ($\gamma \approx$ 0.5772156649). Finally, the anomaly score $s$ of a sample $x$ is calculated as follows \cite{Liu2008isolation}:
\begin{equation}
    s(x,\psi) = 2^{-\frac{E(h(x))}{c(\psi)}}
\end{equation}

Anomalies located within low-density regions typically experience early isolation, leading to shorter average path lengths. Consequently, the anomaly scores assigned to these anomalies tend to be higher than those assigned to normal samples. In other words, within the range of $0<s\leq 1$, instances exhibiting anomaly scores approaching 1 are more likely to represent anomalies.

\section{Methodology}\label{sec:method}
In this section, we elucidate the strategies implemented by RiForest to enhance the performance of iForest, systematically delineating each step. First, RiForest generates a candidate set of split features that combines existing attributes with randomly projected hyperplanes, leveraging the distinct advantages of both feature types for splitting. Next, RiForest employs dimension entropy to select a split feature from this set that effectively separates anomalies from normal samples. To further improve splitting efficacy, the split point on the selected feature is determined using the valley emphasis method. Finally, to increase the divergence in anomaly scores between normal samples and anomalies, RiForest computes anomaly scores by employing varied path lengths for each branch. This is accomplished by assessing the variation in the proportion of samples assigned to the left and right child nodes at each branch, serving as an indicator of the branch's significance from an anomaly detection perspective. Table~\ref{tab:comparison} provides an overview of the different design choices adopted by RiForest compared to other iForest-based methods.
\begin{table}[!tb]
\centering
\caption{Comparison between iForest based methods} \label{tab:comparison}
\footnotesize
\begin{tabular}{m{1.6cm} m{2.33cm} m{2.42cm} m{2.42cm} m{2.9cm}}
\hline
Methods                        & Candidate \newline hyperplane set                               & Hyperplane \newline selection               & Split point \newline selection                         & Path length assignment                             \\ \hline
iForest\cite{Liu2008isolation}                 & Original features                                        & Uniform at random               & Uniform at random                                         & Fixed to 1                                          \\ 
SCiForest\cite{Liu2010on}               & Projected features                              & Deterministic                      & Standard deviation gain         & Varying depending on \newline acceptable range               \\ 
OiForest\cite{Liu2018an}                & Original features                                        & Deterministic                      & Distance between\newline means    & Fixed to 1                                          \\ 
EiForest\cite{Liao2019entropy}                & Original features                                        & Non-uniform \newline at random           & Mid of blank space                            & Varying depending on \newline blank space size               \\ 
EXiForest\cite{Hariri2021extended}               & Projected features                              & Uniform at random               & Uniform at random                                         & Fixed to 1                                          \\ 
GiForest\cite{Lesouple2021generalized}                & Projected features                              & Uniform at random               & Uniform at random                                         & Fixed to 1                                          \\ 
SSPiForest\cite{Tan2022sparse}              & Projected features                              & Uniform at random               & Uniform at random                                         & Fixed to 1                                          \\ 
DiForest\cite{xu2023deep}                & Features of neural \newline representations                              & Uniform at random               & Uniform at random                                         & Fixed to 1                                          \\ \hline
RiForest (ours) & Original features + \newline Projected features & Non-uniform \newline at random  & Valley emphasis \newline method               & Varying depending on \newline proportion difference \\
\hline
\end{tabular}
\end{table}

\subsection{Generation of a candidate hyperplane set}
In most algorithms, the selection of split features is typically limited to either existing attributes or hyperplanes generated through random projection. However, RiForest takes a dual approach by simultaneously considering both existing attributes and hyperplanes generated through random projection. In this context, the existing attributes can be perceived as hyperplanes represented by unit vectors. For instance, the $i$-th feature can be conceptualized as a hyperplane represented by $\vec{e_i}$, where the $i$-th element equals 1 while all others are set to zero. Consequently, RiForest enriches the hyperplane set, which initially consists of existing attributes, by introducing an additional $\tau$ hyperplanes generated through random projection. This comprehensive hyperplane ensemble forms the basis for further processing.

To generate the random hyperplanes, RiForest adopts the soft sparse random projection \cite{Tan2022sparse} from various methodologies \cite{Liu2010on,Hariri2021extended,Lesouple2021generalized,Tan2022sparse}, which is a modification of sparse random projection \cite{Achlioptas2003Database}. In this technique, the elements of the projected vectors can be set to 0 based on a user-defined probability. This approach reduces the influence of unrelated variables and produces hyperplanes with directions that are challenging to create using other methods that consider all variables in the projection. By utilizing soft sparse random projection, RiForest can identify effective hyperplanes for anomaly branching from diverse angles \cite{Tan2022sparse}. The $i$th element of each random projection vector, $r_i$ is determined as follows:
\begin{equation}
    r_i = \sqrt{3s}\begin{cases}
          U(0,1) & \textrm{with probability } \frac{1}{2s} \\
          0 & \textrm{with probability } 1-\frac{1}{s} \\
          U(-1,0) & \textrm{with probability } \frac{1}{2s}
          \end{cases}
\end{equation}
where $U(x,y)$ denotes the uniform distribution between $x$ and $y$, and $s\in (+1,\infty]$ is a hyperparameter to control the sparsity, $\lambda=1-\frac{1}{s} \in (0,1)$ of the random projection vectors. 

However, determining the optimal sparsity for different datasets in advance is challenging, and finding the optimal sparsity typically requires labeled data. To address these issues and enhance the diversity of trees, the proposed algorithm randomly sets the sparsity of projection vectors for each tree. Specifically, for each tree, the sparsity $\lambda$ is randomly selected from a uniform distribution between 0 and 1, aiming to mitigate these challenges and promote tree diversity.

\subsection{Determination of the split feature}
Next, based on the candidate hyperplane set for branching, RiForest employs a unique approach distinct from the random feature selection method commonly used for split generation. Instead, it adopts a strategy similar to that of EiForest \cite{Liao2019entropy}, focusing on selecting features that effectively separate anomalies from normal samples. Specifically, RiForest utilizes dimension entropy to eliminate hyperplanes that are irrelevant to anomaly detection from the candidate hyperplane set as same as EiForest.

Dimension entropy quantifies the uniformity of the one-dimensional sample distribution. To compute the dimension entropy for a given hyperplane $i$, the projected data points onto that hyperplane are initially divided into $L$ segments of equal distance. Subsequently, the dimension entropy, denoted as $ent_{i}$, is calculated using the following equation \cite{Liao2019entropy}:
\begin{equation}
ent_{i} = -\sum_{j=1}^{L} p_j \log p_j
\end{equation}
Here, $p_j$ represents the probability of the $j$-th bin, computed as $p_j = n_j/n$, where $n_j$ stands for the number of samples in the $j$-th bin and $n$ represents the total number of samples. A higher even distribution of samples across all bins results in a larger value for $ent_{i}$. Consequently, hyperplanes with smaller $ent_{i}$ suggest uneven sample distribution, indicating a greater potential for effectively distinguishing between normal samples and anomalies.

Given that the scale of $ent_{i}$ may vary depending on the chosen $L$, it is normalized by dividing it by the maximum dimension entropy value corresponding to $L$, effectively transforming it into a value between 0 and 1. Following this transformation, if hyperplanes exist with dimension entropy values smaller than $\alpha$, one of them is chosen at random. In such cases, the split point is determined using the valley emphasis method described in Section~\ref{sec:valley}. If no hyperplanes meet this criterion, a hyperplane is randomly selected from the entire hyperplane set, and the split point is positioned at the midpoint of the projected distribution.

\subsection{Determination of the split point}\label{sec:valley}
After selecting the split feature, the next step is to determine a split point on the chosen feature. To create effective splits for anomaly detection, it is essential for the split points to efficiently distinguish anomaly samples from normal ones. Identifying valley points is crucial, as they represent areas with minimal overlap between the distributions of normal and anomaly samples. To achieve this, RiForest incorporates the valley emphasis method, widely used in image segmentation.

Thresholding is a technique used in image segmentation to divide an image into distinct regions based on pixel intensity levels. It separates objects of interest from the background by identifying valley points—where histogram intensity has minimal overlap between modes in the gray-level histogram \cite{Ng2006automatic,Al2010image}. The most commonly used thresholding method, Otsu method \cite{Otsu1975a}, finds a globally optimal threshold by maximizing variance between modes \cite{yuheng2017image}. Several iForest-based methods \cite{Liu2010on,Liu2018an} adopt similar principles for determining split points. However, while Otsu method performs well with bimodal distributions of comparable sample sizes, it struggles to find precise boundaries between two modes when the distribution is not clearly bimodal or when there is a significant difference in mode sizes, often setting the peak of a unimodal distribution as the threshold. The valley emphasis method improves on this limitation of Otsu method by enabling accurate threshold detection even in distributions similar to unimodal ones. By assigning higher weights to lower bins, the valley emphasis method guides the threshold toward valley points between two modes or the bottom rim of a unimodal distribution.

In real-world datasets, anomalies are typically far fewer in number than normal samples, and they may be scattered, often resulting in an indistinct separation between anomaly and normal distributions. In such cases, the distribution of features resembles a unimodal distribution, and the valley emphasis method, by setting the split point at the bottom rim of the normal distribution, can more effectively delineate the boundaries of normal samples compared to other image segmentation techniques or prior splitting criteria used in iForests \cite{Liu2018an,Liu2010on,Liao2019entropy,Otsu1975a}. Consequently the valley emphasis method is anticipated to demonstrate enhanced performance across a broader spectrum of anomaly datasets with varying characteristics.

To determine the optimal split point, RiForest utilizes the bins used for calculation of the dimension entropy. In this setting, the valley emphasis method identifies a bin with minimal height while maximizing the variance difference between the divided left and right regions, and then employs the upper bound of this bin as the split point. The objective function to find optimal threshold $t^{*}$ of the valley emphasis method \cite{Ng2006automatic} can be applied for determining the optimal split point as follows:
\begin{equation}
    t^{*} = \arg\max_{ 1<t<L} \{(1-p_t)(w_{L}(t)\mu_{L}^2(t)+w_{R}(t)\mu_{R}^2(t))\} \label{eq:valley}
\end{equation}
where $w_{L}(t)$ and $w_{R}(t)$ denotes the probabilities of the left and right child nodes, which can be computed as follows:
\begin{equation}
    w_{L}(t) = \sum_{j=1}^{t} p_j, \textrm{   } w_{R}(t) = \sum_{j=t+1}^{L} p_j
\end{equation}
In addition, $\mu_{L}(t)$ and $\mu_{R}(t)$ denotes the mean values of the left and right child nodes, which can be computed as follows:
\begin{equation}
    \mu_{L}(t) = \sum_{j=1}^{t} jp_j/w_{L}(t),\textrm{   } \mu_{R}(t) = \sum_{j=t+1}^{L} jp_j/w_{R}(t)
\end{equation}
In Eq.~\eqref{eq:valley}, the first term, $1-p_t$ is the weight and the second term $w_{L}(t)\mu_{L}^2(t)+w_{R}(t)\mu_{R}^2(t)$ is the between-class variance of the child nodes. As the probability of bin $t$ decreases and the between-class variance between two child nodes increases, the value of the objective function increases.

\subsection{Calculation of anomaly scores}
Unlike the original iForest, RiForest assigns varying path lengths to different branches to better distinguish anomaly scores between normal samples and anomalies. The core of this approach is assigning shorter path lengths to splits that are more effective for anomaly detection, particularly those that clearly separate a minority of anomalies from normal samples.

The path length $pl$ is defined based on the difference in sample proportions between the left and right child nodes as follows:
\begin{equation}
    pl = 1 - \left|\sum_{j\in C_l} p_j - \sum_{j\in C_r} p_j \right|
\end{equation}
where $C_l$ and $C_r$ denote the bins allocated to the left and right child nodes, respectively. This path length calculation is exclusively applied to splits derived from the valley emphasis method. When the split is based on the midpoint of a randomly selected hyperplane from the entire hyperplane set, the path length is set to 1, as such splits lack informative value for distribution separation.

Algorithm~\ref{alg:tree} details the complete process of generating a single tree in RiForest as described in this section.

\begin{algorithm}
\caption{Create a single tree, $RiTree$ in RiForest}\label{alg:tree}
\textbf{Inputs:}
\textnormal{$X$ - input data, $\psi$ - current tree height, $L$ - number of bins, $\alpha$ - dimension entropy threshold, $\tau$ - number of random hyperplanes}\\
\textbf{Output:}
\textnormal{an $RiTree$}
\begin{algorithmic}[1]
\Function{RiTree}{$X_s,e,l,L,\alpha,\tau,\lambda$}:
    \If{$e \geq l$ or $|X_s| \leq 1$} 
        \State $Node.Size \gets |X_s|$, $Node.Type \gets$ `ext'
    \Else
        \State $H \gets \{\vec{e_1},...,\vec{e_d}\}$
        \Comment{Hyperplane set containing unit vectors.}
        \State $H \gets H \cup \{\vec{r_1},...,\vec{r_\tau}\}$
        \Comment{Random vectors with $\lambda$ sparsity.}
        \State $H' \gets \{\vec{v} \in H | ent(X_s \cdot \vec{v})/\log L < \alpha \}$
        \Comment{Dimension entropy.}
        \If{$H'$ is not empty}
            \State $\vec{p} \gets$ \Call{RandomSelect}{$H'$}
            \Comment{Split feature vector.}
            \State $q \gets$ \Call{ValleyEmphasis}{$X_s \cdot \vec{p}$}
            \Comment{Split point.}
            \State $X_l \gets$ \Call{filter}{$X_s \cdot \vec{p} \leq q$}, $X_r \gets$ \Call{filter}{$X_s \cdot \vec{p} > q$}
            \State $Node.Splitfeature \gets \vec{p}$, $Node.Splitpoint \gets q$
            \State $Node.Pathlength \gets pl$, $Node.Type \gets$ `int'
            \State $Node.Left \gets$ \Call{RiTree}{$X_l,e+1,l,L,\alpha,\tau,\lambda$}
            \State $Node.Right \gets$ \Call{RiTree}{$X_r,e+1,l,L,\alpha,\tau,\lambda$}
        \Else
            \State $\vec{p} \gets$ \Call{RandomSelect}{$H$}
            \Comment{Split feature vector.}
            \State $q \gets$ \Call{Midpoint}{$X_s \cdot \vec{p}$}
            \Comment{Split point.}
            \State $X_l \gets$ \Call{filter}{$X_s \cdot \vec{p} \leq q$}, $X_r \gets$ \Call{filter}{$X_s \cdot \vec{p} > q$}
            \State $Node.Splitfeature \gets \vec{p}$, $Node.Splitpoint \gets q$
            \State $Node.Pathlength \gets 1$, $Node.Type \gets$ `int'
            \State $Node.Left \gets$ \Call{RiTree}{$X_l,e+1,l,L,\alpha,\tau,\lambda$}
            \State $Node.Right \gets$ \Call{RiTree}{$X_r,e+1,l,L,\alpha,\tau,\lambda$}
        \EndIf
    \EndIf
    \State \Return $Node$
\EndFunction

\Function{Main}{}:
    \State $e \gets 0$, $l \gets$ ceil$(\log_2 \psi)$  
    \Comment{Current tree height and height limit.}
    \State $X_s \gets$ \Call{Sample}{$X,\psi$}
    \State draw $\lambda \sim U(0,1)$
    \Comment{Sparsity of random vector.}
    \State \Return \Call{RiTree}{$X_s,e,l,L,\alpha,\tau,\lambda$}
\EndFunction
\end{algorithmic}
\end{algorithm}

\section{Experimental framework}\label{sec:experiment}
\subsection{Data}
In this study, we assessed various outlier detection algorithms, including RiForest, across 24 benchmark datasets that are commonly utilized in prior research. Among these benchmark datasets, the Creditcard dataset was sourced from Kaggle\footnote{https://www.kaggle.com/datasets/mlg-ulb/creditcardfraud}, and the Aloi dataset was retrieved from the Unsupervised Anomaly Detection Dataverse in the Harvard Dataverse\footnote{https://doi.org/10.7910/DVN/OPQMVF}. The remaining datasets were obtained from the Outlier Detection DataSets\footnote{https://odds.cs.stonybrook.edu}. The Unsupervised Anomaly Detection Dataverse and Outlier Detection DataSets offer datasets collected from diverse origins, which have been processed and formatted suitably for anomaly detection purposes. Details on each dataset's original source and preprocessing methods can be found on the respective websites where the datasets were obtained. Table~\ref{tab:data} summarizes the dataset names, sample and feature counts, as well as anomaly rates.

\begin{table}[!tb]
\centering
\caption{Detailed information for 24 benchmark datasets} \label{tab:data}
\footnotesize
\begin{tabular}{@{}lrrr@{}}
\toprule
            & \# of samples & \# of variables & Anomaly rate (\%) \\ \midrule
Smtp        & 95,156             & 3                  & 0.03       \\
Creditcard  & 284,807            & 29                 & 0.17       \\
Http        & 567,479            & 3                  & 0.40       \\
Forestcover & 286,048            & 10                 & 0.90       \\
Satimage-2  & 5,803              & 36                 & 1.20       \\
Speech      & 3,686              & 400                & 1.65       \\
Pendigits   & 6,870              & 16                 & 2.27       \\
Mammography & 11,183             & 6                  & 2.32       \\
Thyroid     & 3,772              & 6                  & 2.50       \\
Optdigits   & 5,216              & 64                 & 3.00       \\
Aloi        & 50,000             & 27                 & 3.02       \\
Musk        & 3,062              & 166                & 3.20       \\
Vowels      & 1,456              & 12                 & 3.40       \\
Glass       & 214               & 9                  & 4.20       \\
Letter      & 1,600              & 32                 & 6.25      \\
Shuttle     & 49,097             & 9                  & 7.00       \\
Annthyroid  & 7,200              & 6                  & 7.42       \\
Wine        & 129               & 13                 & 7.70       \\
Cardio      & 1,831              & 21                 & 9.60       \\
Vertebral   & 240               & 6                  & 12.50      \\
Satellite   & 6,435              & 36                 & 32.00      \\
Pima        & 768               & 8                  & 35.00      \\
Breastw     & 683               & 9                  & 35.00      \\
Ionosphere  & 351               & 33                 & 36.00      \\ \bottomrule
\end{tabular}
\end{table}

\subsection{Experimental design}
In this study, the performance of RiForest was compared with various iForest-based algorithms, including iForest \cite{Liu2008isolation}, EiForest \cite{Liao2019entropy}, GiForest \cite{Lesouple2021generalized}, SCiForest \cite{Liu2010on}, SSPiForest \cite{Tan2022sparse}, and DiForest \cite{xu2023deep}, as well as probabilistic algorithms such as Copula-Based Outlier Detector (COPOD) \cite{li2020copod} and  Empirical Cumulative Distribution Functions for Outlier Detection (ECOD) \cite{li2022ecod}. These algorithms exhibit various characteristics, allowing for a comprehensive performance comparison. 

The experiments were designed to evaluate three aspects: outlier detection performance, stability, and robustness to noisy variables. iForest-based methods involve various random factors, leading to different outcomes with each run on the same dataset. Therefore, to assess outlier detection performance and stability, 20 experiments were conducted for each algorithm on individual datasets, and the results were aggregated for iForest-based methods. Since COPOD and ECOD do not have randomness, experiments were conducted only once per dataset. The robustness to noisy variables was evaluated following the experimental setup used in \cite{Liao2019entropy} and \cite{Tan2022sparse}. In these experiments, outlier detection performance was measured by gradually increasing the number of randomly generated noisy variables (ranging from 10 to 100), which followed a Gaussian distribution. This allowed us to observe how performance changed with varying numbers of noisy variables. Six datasets---Satimage-2, Ionosphere, Glass, Letter, Pima, and Thyroid---were selected for these experiments considering their anomaly rates and feature counts.

For RiForest, GiForest, and SSPiForest, which generate hyperplanes using random projection, experiments were conducted after standardizing the experimental datasets so that the mean and variance of each variable were 0 and 1, respectively. Following the authors' code, min-max normalization was applied to each variable for DiForest, while standardization was performed for COPOD and ECOD.

In most real-world scenarios where labeled data is limited, optimizing hyperparameter settings for each dataset is challenging. Therefore, in this study, instead of using individually optimized hyperparameters for each dataset, we compared the performance of algorithms based on default hyperparameter settings provided in the literature or settings that have demonstrated good performance on multiple datasets.

First, for the common hyperparameters of all iForest-based algorithms, except for DiForest, the number of trees and the subsample size were set to 100 and 256, respectively, following the study that originally proposed iFores \cite{Liu2008isolation}. For DiForest, 50 representations and 6 iTrees per representation were trained, while the subsample size was set the same as the other iForest-based algorithms \cite{xu2023deep}. For EiForest, the number of bin segments ($L$) was set to 10, and $\alpha$ was set to 0.8 \cite{Liao2019entropy}. For SCiForest, the number of randomly generated hyperplanes was set to 10, and the number of variables used in hyperplane generation was set to 2 \cite{Liu2010on}. In the case of SSPiForest, the optimal values for the number of random hyperplanes and sparsity of the projection vector were not provided for individual datasets, even though the ranges of values for these parameters explored in the experiments were provided in \cite{Tan2022sparse}. Hence, in this study, the best-performing values were determined through experimentation on average. Ultimately, the values presented in the experimental results are based on the number of hyperplanes equal to $\lceil d^{3/2}\rceil$ and a sparsity of 0.8. Similarly, for RiForest, the number of bin segments ($L$) was set to 10, $\alpha$ was set to 0.8, and the number of random hyperplanes added to the hyperplane set ($\tau$) was determined to be 5 through experimentation. Details of our hyperparameter analysis can be found in Section~\ref{sec:hyper}.

\subsection{Evaluation metrics}
The primary evaluation metric employed to assess anomaly detection performance is the Area Under the Curve of the Receiver Operating Characteristics (AUROC), a widely used measure in various anomaly detection studies. The AUROC values are presented as the average results from 20 experiments, except for COPOD and ECOD.

In addition, given the inherent challenge of finding optimal hyperparameter settings, an evaluation was conducted to gauge algorithm stability across datasets, regardless of specific dataset characteristics. To achieve this, the improvement rate was computed for each dataset based on the average performance. The improvement rate (\%) for algorithm $m$ on dataset $D$, $IR(m,D)$ is defined as follows:
\begin{equation}
IR(m,D) = \frac{AUROC(m,D)-\overline{AUROC}(D)}{\overline{AUROC}(D)} \times 100
\label{eq:IR}
\end{equation}
Here, $m$ represents the specific anomaly detection algorithm, $D$ signifies the particular dataset, $AUROC(m,D)$ stands for the average AUROC value of algorithm $m$ for dataset $D$ across 20 repetitions, and $\overline{AUROC}(D)$ represents the average AUROC values of different algorithms for dataset $D$. 

To assess the stability of anomaly detection, the coefficient of variation ($CV$) of AUROC values was calculated across the 20 repetitions. Since COPOD and ECOD were each experimented on only once per dataset, CV measurement was not performed.

Moreover, to compare the algorithms holistically based on overall average performance rather than individual datasets, the mean values of AUROC, improvement rate, and the coefficient of variation were computed. Additionally, the average performance ranking for each algorithm across all datasets was calculated.

\section{Results}\label{sec:result}
\subsection{Anomaly detection performance and stability}
Tables~\ref{tab:AUROC} and~\ref{tab:IR} provide the results on AUROC values and improvement rates for each dataset and algorithm. The rows, labeled as ``Average'' and ``Rank'' in Table~\ref{tab:AUROC} represent the average values and ranks of AUROC calculated across the 24 datasets for each algorithm. Additionally, ``Count'' indicates the number of datasets where each algorithm performed best. Notably, in Table~\ref{tab:IR}, we've omitted Rank and Count, as they aligned with the average ranks and the number of best-performed datasets obtained by AUROC in Table~\ref{tab:AUROC}. The best values for each dataset are highlighted in bold for both Tables~\ref{tab:AUROC} and~\ref{tab:IR}.

\begin{table}[!tb]
\footnotesize
\centering
\caption{Experimental results: $AUROC$} \label{tab:AUROC}
\setlength{\tabcolsep}{4pt}
\begin{tabular}{lS[table-format=1.4]S[table-format=1.4]S[table-format=1.4]S[table-format=1.4]S[table-format=1.4]S[table-format=1.4]S[table-format=1.4]S[table-format=1.4]S[table-format=1.4]}
\toprule
\multicolumn{1}{l}{Data} & \multicolumn{1}{c}{RiForest} & \multicolumn{1}{c}{iForest} & \multicolumn{1}{c}{EiForest} & \multicolumn{1}{c}{GiForest}  & \multicolumn{1}{c}{SCiForest} & \multicolumn{1}{c}{SSPiForest} & \multicolumn{1}{c}{DiForest} & \multicolumn{1}{c}{COPOD} & \multicolumn{1}{c}{ECOD} \\ \midrule
Smtp & 0.9339 & 0.9083 & \textbf{0.9413} & 0.8960  & 0.9117 & 0.8936 & 0.7161 & 0.9120 & 0.8801\\
Creditcard & 0.9496 & 0.9501 & 0.9475 & \textbf{0.9519} & 0.9278 & 0.9515 & 0.9499 & 0.9475 & 0.9496 \\
Http & 0.9977 & \textbf{0.9995} & 0.9973 & 0.9970 & 0.9994 & 0.9950 & 0.9933 & 0.9915 & 0.9786 \\
Forestcover & 0.8680 & 0.8742 & 0.9058 & 0.9400 & 0.6762 & 0.9109 & \textbf{0.9625} & 0.8841 & 0.9204 \\
Satimage-2 & \textbf{0.9985} & 0.9936 & \textbf{0.9985} & 0.9957 & 0.9947 & 0.9950 & 0.9973 & 0.9745 & 0.9649 \\
Speech & 0.4861 & 0.4782 & 0.4780 & 0.4763 & 0.4868 & 0.4740 & 0.4659 & \textbf{0.4911} & 0.4697\\
Pendigits  & 0.9659 & 0.9444 & \textbf{0.9736} & 0.9322 & 0.9547 & 0.9395 & 0.9510 & 0.9048 & 0.9274 \\
Mammography & 0.8016 & 0.8605 & 0.7892 & 0.8681 & 0.5254 & 0.8702 & 0.7794 & 0.9053 & \textbf{0.9062} \\
Thyroid & 0.9767 & \textbf{0.9787} & 0.9602 & 0.9566 & 0.9714 & 0.9702 & 0.9616 & 0.9393 & 0.9771 \\
Optdigits & \textbf{0.7963} & 0.6973 & 0.6875 & 0.5288 & 0.7774 & 0.5402 & 0.5687 & 0.6824 & 0.6045 \\
Aloi & 0.5436 & 0.5401 & 0.5458 & \textbf{0.5552} & 0.5268 & 0.5489 & 0.5487 & 0.5143 & 0.5289 \\
Musk & \textbf{1.0000} & 0.9997 & \textbf{1.0000}  & 0.9995 & 0.9998 & 0.9996 & 0.9782 & 0.9463 & 0.9559  \\
Vowels & \textbf{0.9061} & 0.7558 & 0.8973 & 0.7830 & 0.8839 & 0.7625 & 0.7957 & 0.4958 & 0.5929  \\
Glass & 0.7086 & 0.6952 & 0.6039 & 0.6978 & 0.7743 & 0.7007 & \textbf{0.7896} & 0.6450 & 0.6206 \\
Letter & \textbf{0.6818} & 0.6267 & 0.5047 & 0.6173 & 0.5067 & 0.6112 & 0.6437 & 0.5596 & 0.5723  \\
Shuttle & 0.9967 & 0.9971 & \textbf{0.9989} & 0.9920 & 0.9987 & 0.9958 & 0.9668 & 0.9945 & 0.9929 \\
Annthyroid & \textbf{0.9128} & 0.8149 & 0.8516 & 0.6842 & 0.8092 & 0.7589 & 0.6764 & 0.7760 & 0.7887  \\
Wine & \textbf{0.8917} & 0.7742 & 0.8880 & 0.7620 & 0.8800 & 0.7961 & 0.4003 & 0.8672 & 0.7328 \\
Cardio & 0.8554 & 0.9234 & 0.9125 & 0.9335 & 0.8805 & 0.9304 & 0.9280 & 0.9219 & \textbf{0.9350} \\
Vertebral & 0.2716 & 0.3620 & 0.2613 & 0.3779 & \textbf{0.6954} & 0.3673 & 0.4967 & 0.3349 & 0.4200 \\
Satellite & \textbf{0.8506} & 0.7077 & 0.8429 & 0.6818 & 0.5732 & 0.6830 & 0.7411 & 0.6335 & 0.5830 \\
Pima & \textbf{0.6852} & 0.6765 & 0.6278 & 0.6739 & 0.6387 & 0.6765 & 0.6190 & 0.6540 & 0.5944 \\
Breastw & 0.9685 & 0.9863 & 0.7965 & 0.9773 & 0.9899 & 0.9865 & 0.7584 & \textbf{0.9944} & 0.9914 \\
Ionosphere & 0.9090 & 0.8554 & \textbf{0.9335} & 0.8652 & 0.8934 & 0.8643 & 0.9063 & 0.7990 & 0.7353 \\ \midrule
Average & \textbf{0.8315} & 0.8083 & 0.8060 & 0.7976 & 0.8032 & 0.8009 & 0.7748 & 0.7820 & 0.7759  \\
Rank  & \textbf{3.2917} & 4.5000 & 4.4167 & 5.1250 & 4.7500 & 4.8750 & 5.4583 & 6.2083 & 6.1667 \\ 
Count  & \textbf{9} & \textrm{2} & \textrm{6} & \textrm{2} & \textrm{1} & \textrm{0} & \textrm{2} & \textrm{2} & \textrm{2} \\ \bottomrule
\end{tabular}%
\end{table}

\begin{table}[!tb]
\centering
\footnotesize
\caption{Experimental result: $IR$ (\%)} \label{tab:IR}
\setlength{\tabcolsep}{4pt}
\begin{tabular}{lS[detect-weight,mode=text,table-format=-1.4]S[detect-weight,mode=text,table-format=-2.4]S[detect-weight,mode=text,table-format=2.4]S[detect-weight,mode=text,table-format=2.4]S[detect-weight,mode=text,table-format=2.4]S[detect-weight,mode=text,table-format=2.4]S[detect-weight,mode=text,table-format=2.4]S[detect-weight,mode=text,table-format=2.4]S[detect-weight,mode=text,table-format=2.4]}
\toprule
\multicolumn{1}{l}{Data} & \multicolumn{1}{c}{RiForest} & \multicolumn{1}{c}{iForest} & \multicolumn{1}{c}{EiForest} & \multicolumn{1}{c}{GiForest}  & \multicolumn{1}{c}{SCiForest} & \multicolumn{1}{c}{SSPiForest} & \multicolumn{1}{c}{DiForest} & \multicolumn{1}{c}{COPOD} & \multicolumn{1}{c}{ECOD} \\ \midrule
Smtp            & 4.5789 &2.0189 &\B5.3189  & 0.7889 & 2.3589 & 0.5489 & -17.2011 & 2.3889 & -0.8011   \\
Creditcard      & 0.2333  & 0.2833 & 0.0233   &\B0.4633 & -1.9467 & 0.4233 & 0.2633 & 0.0233 & 0.2333   \\
Http            & 0.3333 & \B0.5133 & 0.2933  & 0.2633 & 0.5033 & 0.0633 & -0.1067 & -0.2867 & -1.5767   \\
Forestcover     & -1.4456  & -0.8256 &2.3344 & 5.7544  & -20.6256 & 2.8444 &\B 8.0044 & 0.1644 & 3.7944   \\
Satimage-2      &\B 0.8200 & 0.3300 & \B0.8200 & 0.5400 & 0.4400 &0.4700 & 0.7000 & -1.5800 & -2.5400   \\
Speech          & 0.7556 & -0.0244 & -0.0444 &-0.2144 & 0.8356 & -0.4444 & -1.2544 & \B1.2656 & -0.8744   \\
Pendigits       & 2.2178 &0.0678 & \B2.9878 & -1.1522 & 1.0978 & -0.4222 & 0.7278 & -3.8922 & -1.6322   \\
Mammography     & -1.0167 & 4.8733 & -2.2567  & 5.6333  & -28.6367 & 5.8433 & -3.2367 & 9.3533 & \B9.4433   \\
Thyroid         & 1.0944 & \B1.2944 & -0.5556 & -0.9156 & 0.5644 & 0.4444 & -0.4156 & -2.6456 & 1.1344   \\
Optdigits       &\B 14.2622 & 4.3622 & 3.3822 & -12.4878 & 12.3722 & -11.3478 & -8.4978 & 2.8722 & -4.9178   \\
Aloi            & 0.4467 & 0.0967 &0.6667 & \B1.5967 & -1.2333 & 0.9767 & 0.9567 & -2.4833 & -1.0233  \\
Musk            & \B1.3444 &1.3144 & \B1.3444  & 1.2944 & 1.3244 & 1.3044 & -0.8356 & -4.0256 & -3.0656   \\
Vowels          & \B14.2444 & -0.7956 & 13.3644 & 1.9344 & 12.0244 &-0.1156 & 3.2044 & -26.7856 & -17.0756   \\
Glass           & 1.5744 & 0.2344 & -8.8956 &0.4944 & 8.1444 & 0.7844 & \B9.6744 & -4.7856 & -7.2256   \\
Letter          & \B9.0256  & 3.5156 & -8.6944 & 2.5756 & -8.4844 &1.9656 & 5.2156 & -3.1944 & -1.9244  \\
Shuttle         & 0.4100 &0.4500 & \B0.6300 & -0.0600 & 0.6100 & 0.3200 & -2.5800 & 0.1900 & 0.0300   \\
Annthyroid      & \B12.6944 & 2.9044 & 6.5744 & -10.1656 & 2.3344 &-2.6956 & -10.9456 & -0.9856 & 0.2844 \\
Wine            & \B11.4778 & -0.2722 & 11.1078  & -1.4922 & 10.3078 &1.9178 & -37.6622 & 9.0278 & -4.4122   \\
Cardio          & -5.8000 & 1.0000 &-0.0900 & 2.0100 & -3.2900 & 1.7000 & 1.4600 & 0.8500 & \B2.1600  \\
Vertebral       & -12.6967 & -3.6567 & -13.7267  &-2.0667 & \B29.6833 & -3.1267 & 9.8133 & -6.3667 & 2.1433 \\
Satellite       & \B15.0956 & 0.8056 & 14.3256  & -1.7844 & -12.6444 & -1.6644 &4.1456 & -6.6144 & -11.6644  \\
Pima            & \B3.5644  & 2.6944 & -2.1756 & 2.4344 & -1.0856 & 2.6944 & -3.0556 & 0.4444 & -5.5156   \\
Breastw         & 2.9700  & 4.7500 & -14.2300 & 3.8500 & 5.1100 & 4.7700 & -18.0400 & \B5.5600 & 5.2600   \\
Ionosphere      & 4.6622 &-0.6978 & \B7.1122  & 0.2822 & 3.1022 & 0.1922 & 4.3922 & -6.3378 & -12.7078   \\ \midrule
Average         & \B3.3689 & 1.0515 & 0.8174 & -0.0176 & 0.5361 & 0.3103 & -2.3031 & -1.5768 & -2.1864   \\ \bottomrule

\end{tabular}
\end{table}

\begin{figure}[!tb]
\begin{minipage}{0.5\textwidth}
\begin{center}
\includegraphics[width=\linewidth]{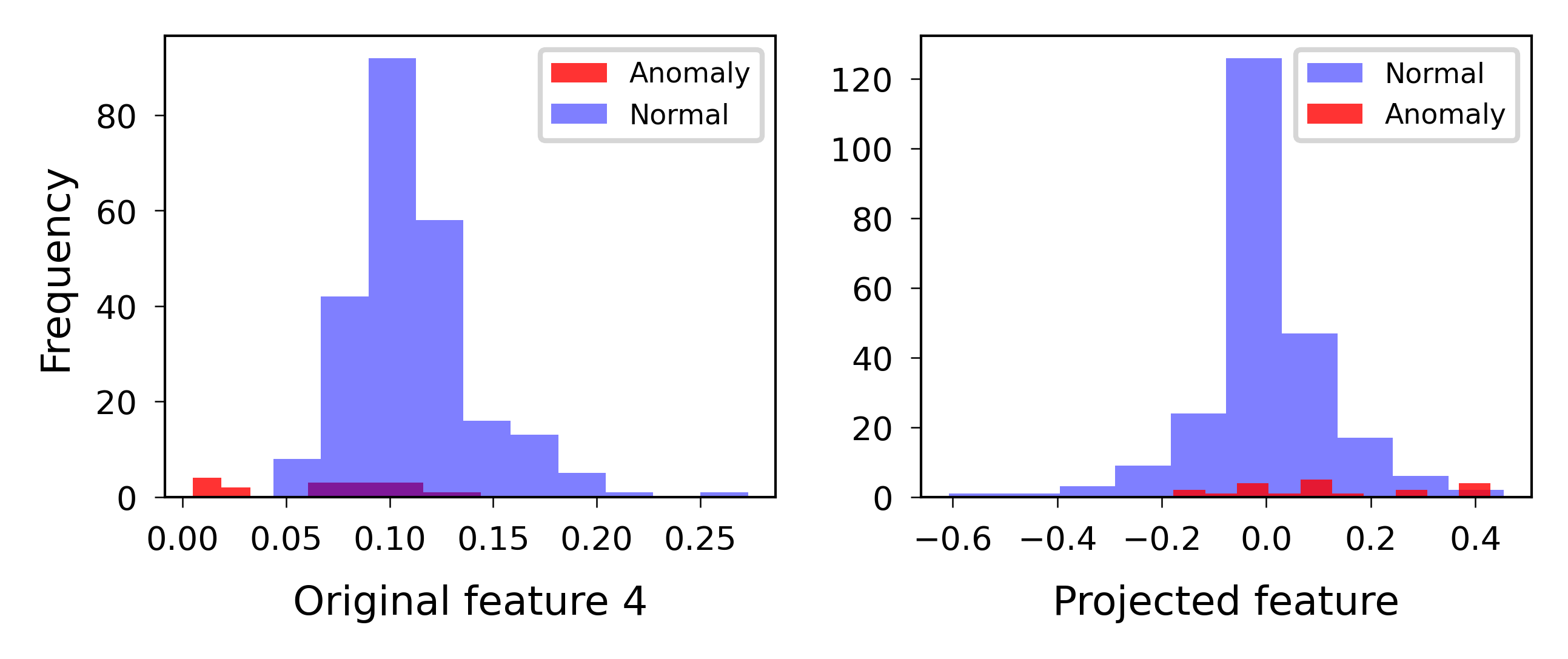} \\
    (a) Annthyroid
\end{center}
\end{minipage}
\begin{minipage}{0.5\textwidth}
\begin{center}
\includegraphics[width=\linewidth]{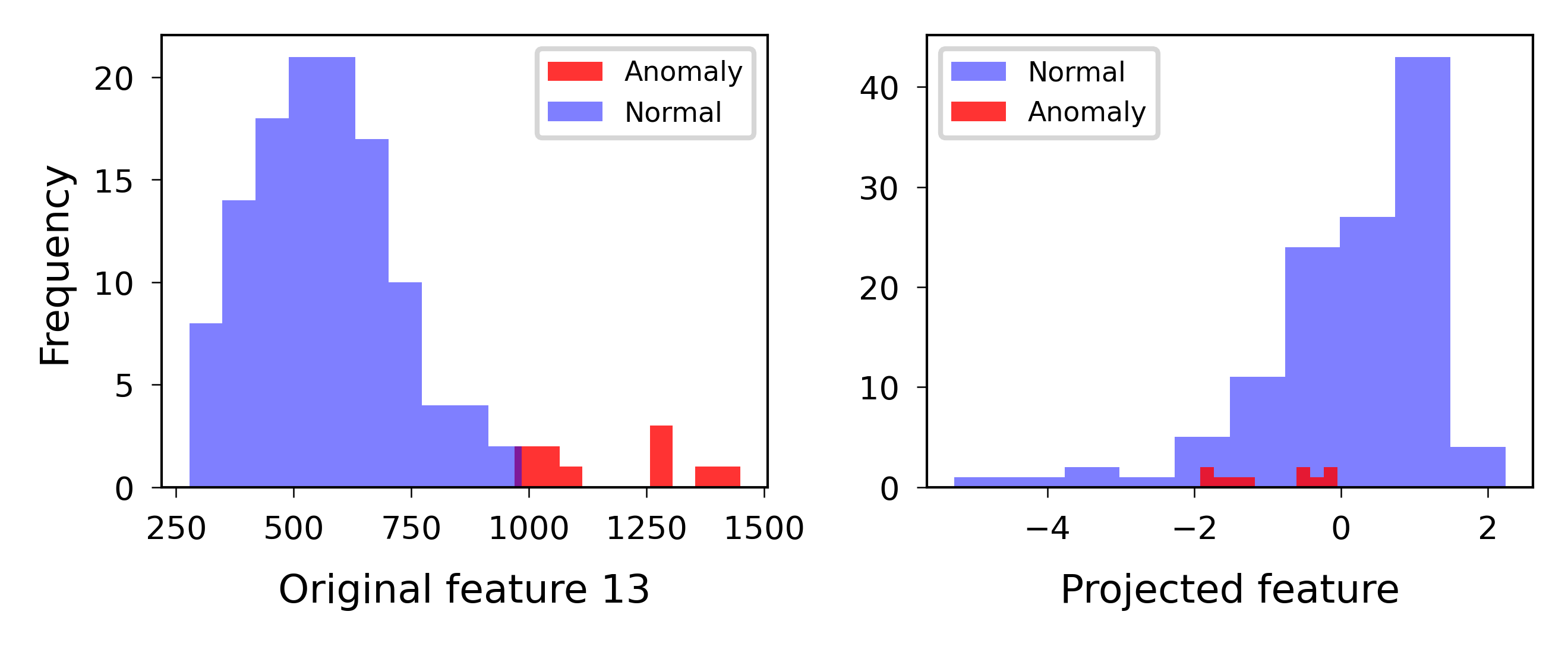} \\
    (b) Wine
\end{center}
\end{minipage}
\begin{minipage}{0.5\textwidth}
\begin{center}
\includegraphics[width=\linewidth]{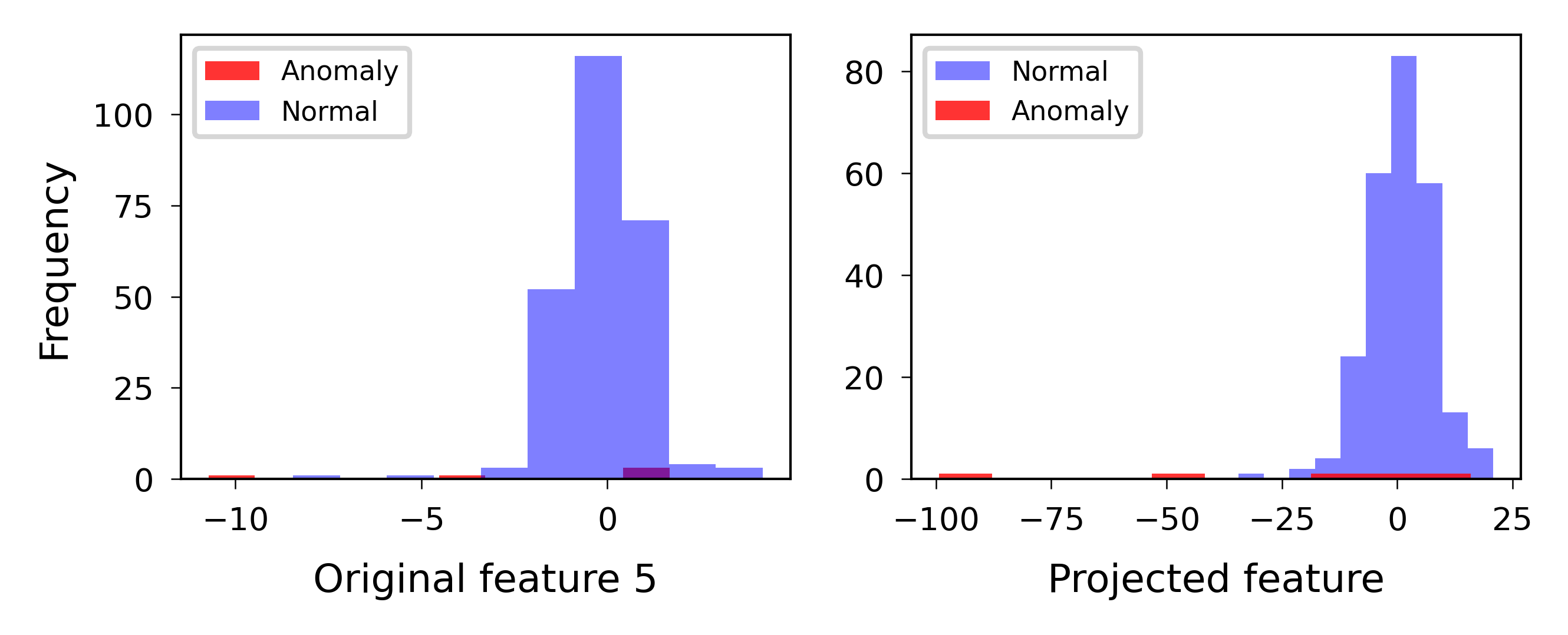} \\
    (c) Creditcard
\end{center}
\end{minipage}
\begin{minipage}{0.5\textwidth}
\begin{center}
\includegraphics[width=\linewidth]{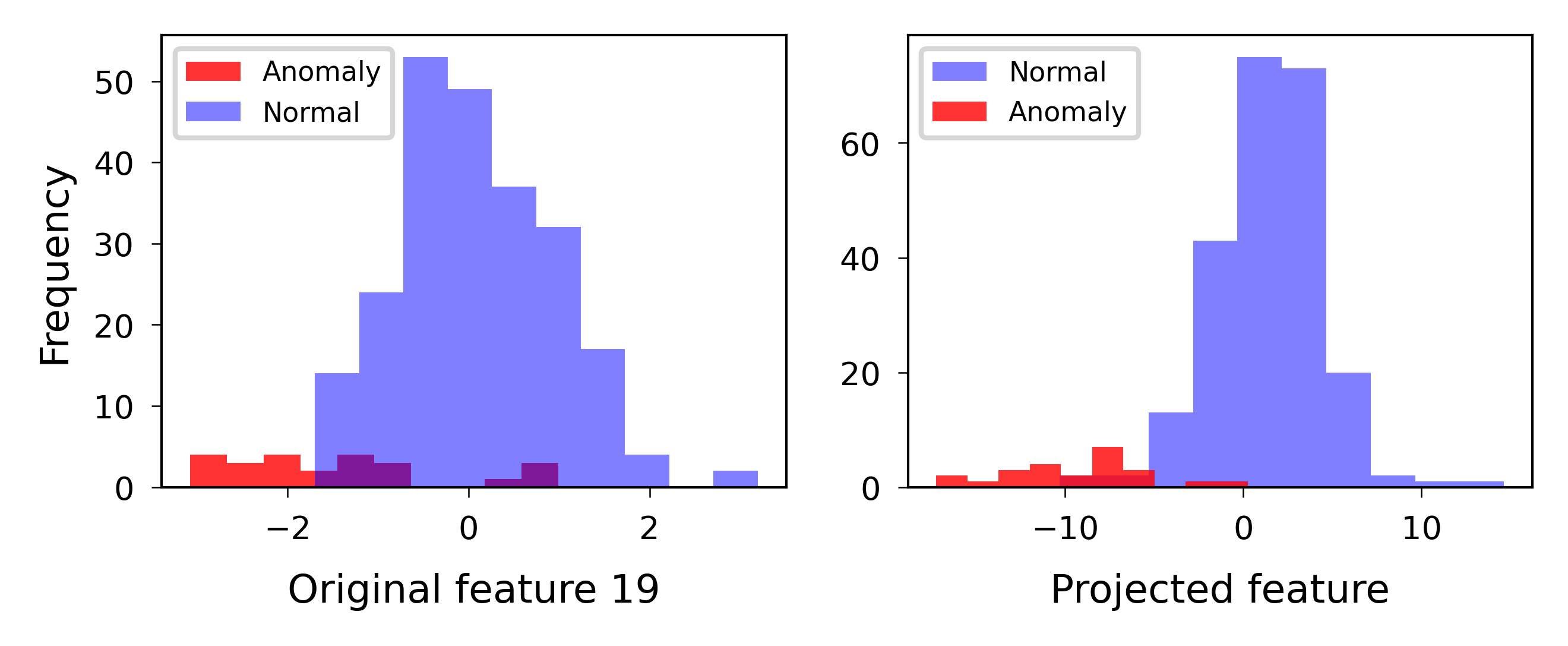} \\
    (d) Cardio
\end{center}
\end{minipage}  
  \caption{Comparison of anomaly separability between original features and random hyperplanes}\label{fig:dist}
\end{figure}

A closer look at Table~\ref{tab:AUROC} reveals that algorithm performance varies depending on the dataset. However, RiForest consistently exhibited the highest AUROC values across 9 out of all datasets, showcasing its superior outlier detection capabilities on a larger subset of datasets compared to other methods. Based on the average AUROC values in Table~\ref{tab:AUROC}, RiForest was followed by iForest, EiForest, SCiForest, SSPiForest, GiForest, COPOD, ECOD and DiForest.

EiForest ranked second to RiForest in both average rank and the number of datasets with top performance, yet showed a lower average AUROC than iForest. These results occur because, in the case of EiForest, it shows excellent anomaly detection performance on certain datasets, but on some others, its performance is significantly worse than that of other iForest-based algorithms. This characteristic is particularly evident in Table~\ref{tab:IR}. While the average AUROC across all datasets in Table~\ref{tab:AUROC} shows little difference between EiForest and iForest, the average IR across all datasets in Table~\ref{tab:IR} reveals that iForest outperforms EiForest by approximately 29\%, highlighting a clear distinction. EiForest achieved high improvement rates on datasets such as Smtp, Pendigits, and Ionosphere. However, on datasets like Glass, Letter, and Breastw, it experienced significant performance degradation, reflected in large negative improvement rates. This decline can be attributed to EiForest's inability to effectively partition the original features, as its largest empty regions fail to separate anomalies in relevant areas. In contrast, RiForest consistently achieved high performance on these datasets by identifying more optimal split points that account for the data distribution. 

SCiForest exhibited the second-highest average AUROC and third-highest average rank among the comparison methods. However, as shown in Table~\ref{tab:IR}, it exhibited significant variability in improvement rates. SCiForest demonstrated very high improvement rates on datasets such as Vertebral and Optdigits but showed substantial negative values on datasets like Forestcover, Mammography, and Satellite. Notably, on datasets where SCiForest exhibited large negative improvement rates, algorithms such as GiForest and SSPiForest---which also perform splits on features projected onto random hyperplanes but select hyperplanes and split points randomly---often achieved relatively high performance. This contrast highlights the limitations of deterministic methods for selecting hyperplanes and split points based on ineffective criteria, particularly when the data distribution is not clearly bimodal.

Among the iForest-based algorithms, SSPiForest achieved the highest average AUROC and rank, followed by GiForest and DiForest. SSPiForest outperformed GiForest on many datasets by mitigating the influence of irrelevant variables through sparse projection. DiForest showed greater performance variability across datasets, resulting in the lowest average AUROC and rank among these algorithms. However, these methods share the common feature of performing random splits on transformed features, either projected onto random hyperplanes or learned through neural networks, rather than using the original features. This may explain the similar performance patterns observed across various datasets.

COPOD and ECOD achieved lower average AUROC and average rank compared to all iForest-based algorithms, except for DiForest, reaffirming the effectiveness of iForest-based approaches. Since COPOD and ECOD detect anomalies based on the tail probabilities of all original variables, their performance is significantly reduced on datasets such as Vowels, Glass, and Satellite, where most variables---apart from a few---exhibit a mixed distribution of anomalies and normal data.

Focusing on Table~\ref{tab:IR} for a comparison of anomaly detection methods in terms of the improvement rate, RiForest exhibits the least performance variation across datasets and achieves a positive improvement rate on most datasets, outperforming the other methods in this regard. RiForest recorded a negative improvement rate on only four datasets—the fewest among the evaluated methods. In comparison, iForest and SSPiForest showed negative improvement rates on six and seven datasets, respectively, while over one-third of the datasets for the remaining methods exhibited negative improvement rates. Notably, DiForest, COPOD, and ECOD displayed negative improvement rates on more than half of the datasets, indicating significantly lower stability than the other methods. These results suggest that while RiForest may not consistently outperform all algorithms on every dataset, it generally demonstrates superior performance across a broad range of datasets.

The stable anomaly detection performance of RiForest's performance, regardless of the dataset's characteristics, can be attributed to its use of both existing features and random hyperplanes for splitting. When comparing algorithm performance across different datasets in Table~\ref{tab:AUROC}, it's apparent that the best-performing algorithm varies depending on the dataset. For instance, in datasets such as Smtp, Pendigits, Annthyroid, Satellite, and Wine, EiForest excels, while GiForest and SSPiForest perform poorly. In contrast, in datasets such as Creditcard, Forestcover, Mammography, Glass, and Cardio, GiForest and SSPiForest shine, with EiForest's performance notably lower. These results suggest that some datasets are better separated into anomalies and normal samples by existing features, while random hyperplanes are more efficient for outlier separation in others. In datasets where RiForest exhibited a negative improvement rate, most projected features displayed a uniform distribution, resulting in a limited number of candidate hyperplanes in the hyperplane set. Furthermore, features projected onto the selected hyperplanes often followed a unimodal distribution in which anomalies and normal data were intermixed. This resulted in ineffective anomaly separation when the split point was positioned at the lower edge of the unimodal distribution.

Figure~\ref{fig:dist} illustrates the distributions of anomalies and normal samples for four different datasets using both one of the original features and a random hyperplane. In datasets like Annthyroid and Wine, where EiForest demonstrated high performance, it is evident that anomalies and normal data are more distinctly separated in the original features. In contrast, in datasets like Creditcard and Cardio, where GiForest and SSPiForest showed better performance, anomalies and normal data are more clearly separated in the projected features. This highlights the varying effectiveness of original features in distinguishing anomalies from normal samples across different datasets. However, in real-world applications, it is challenging to determine which datasets can be effectively handled using only the original features due to the absence of label data. Therefore, to consistently achieve good performance across diverse datasets, it is recommended to utilize both original and projected features, as in RiForest.

\begin{figure}[!tb]
\label{fig:split}
  \centering
  \includegraphics[width=0.9\textwidth]{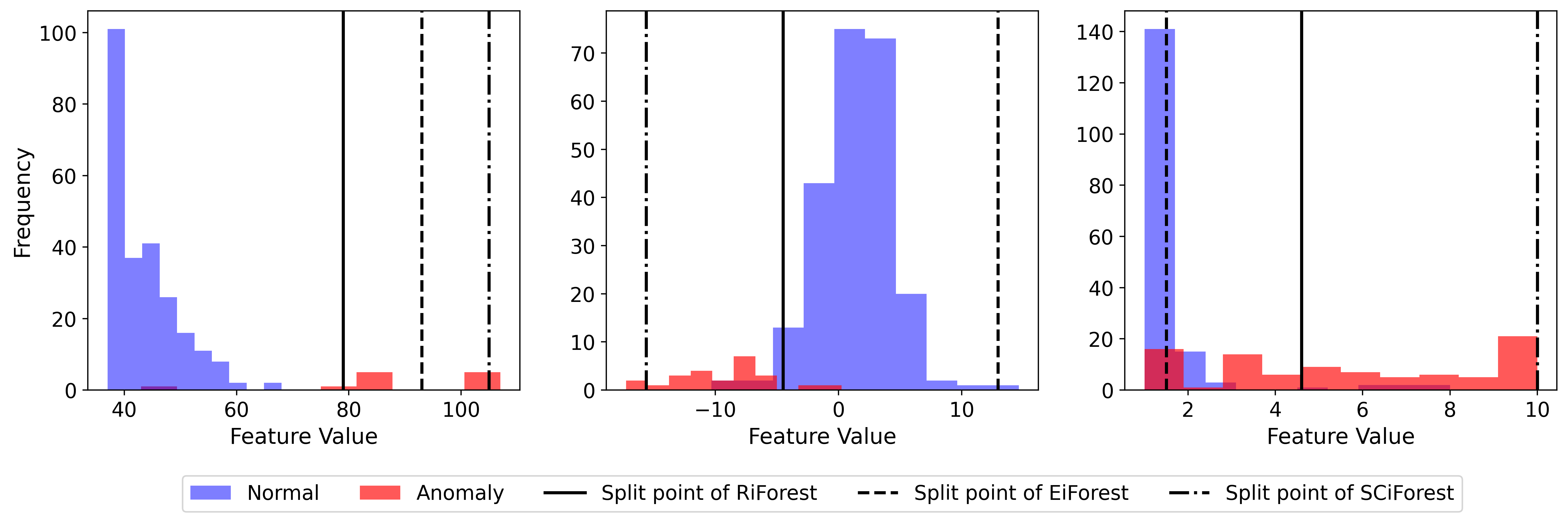}
  \caption{Comparison between the methods to determine a split point}\label{fig:split_point}
\end{figure}

The superior performance of RiForest compared to other algorithms can also be attributed to its use of the valley emphasis method for determining split points. Figure~\ref{fig:split_point} illustrates the positions of split points in the histograms for RiForest, EiForest, and SCiForest across distributions with different characteristics. On the far left, a distribution shows well-separated normal data and anomalies; in the middle, normal and anomaly distributions are merged, forming a single bell-shaped distribution; and on the far right, anomalies appear in a long tail of the normal distribution. In all these distributions, the valley emphasis method consistently selects more optimal split points than other methods. In the left distribution, both EiForest and SCiForest struggle to accurately identify the boundary between anomaly and normal sample distributions, despite the clear separation. Additionally, in the middle and right distributions, SCiForest places split points at the distribution tails, leading to improper anomaly separation. Furthermore, despite SCiForest's higher computational overhead—due to calculating statistics for evaluating separation superiority between each data point—it yielded poorer results than RiForest, which evaluates the values in Eq.~\eqref{eq:valley} for each bin.

Next, Table~\ref{tab:CV} presents the CV values for iForest-based algorithms across 24 datasets. Similar to Table~\ref{tab:AUROC}, this table includes average CV values, average ranks, and the number of datasets where each algorithm recorded the lowest CV values. As with the AUROC results, the algorithm with the lowest CV value varied across datasets. However, RiForest consistently exhibited the lowest CV in 9 out of the 24 datasets, highlighting its overall stability. Additionally, RiForest outperformed other methods in both average CV values and rankings. A notable trend emerges when comparing Tables~\ref{tab:AUROC} and~\ref{tab:CV}. Interestingly, datasets where algorithms achieve high AUROC values often align with datasets where they show low CV values. This suggests that datasets with lower stability may see performance drop-offs in certain repetitions, resulting in reduced AUROC values. 

DiForest and SSPiForest, in particular, exhibited significantly higher average CV values compared to other algorithms. Notably, although DiForest demonstrated the highest stability across the four datasets, it showed the lowest mean CV, indicating that its performance greatly varies depending on the fit of the randomly generated neural network representations. Additionally, the large performance fluctuations observed in SSPiForest can be attributed to the sensitivity of its performance to the suitability of the non-sparsified variable combinations.

\begin{table}[!tb]
\centering
\footnotesize
\caption{Experimental results: $CV$ $(\times 10^{-2})$} \label{tab:CV}
\setlength{\tabcolsep}{4pt}
\begin{tabular}{lS[table-format=1.4]S[table-format=1.4]S[table-format=1.4]S[table-format=1.4]S[table-format=1.4]S[table-format=1.4]S[table-format=1.4]}
\toprule
\multicolumn{1}{l}{Data} & \multicolumn{1}{c}{RiForest} & \multicolumn{1}{c}{iForest} & \multicolumn{1}{c}{EiForest} & \multicolumn{1}{c}{GiForest}  & \multicolumn{1}{c}{SCiForest} & \multicolumn{1}{c}{SSPiForest} & \multicolumn{1}{c}{DiForest} \\ \midrule
Smtp        & 0.5058         & 0.9047          & \textbf{0.1949} & 0.9084          & 0.4878          & 1.0751     & 0.3871          \\
Creditcard & 0.2949          & 0.3101          & 0.3496          & \textbf{0.1651} & 0.3771          & 0.1839     & 0.2940          \\
Http & 0.1574                  & 0.0475 & 0.2079          & 0.1172          & 0.0732          & 0.0786     & \textbf{0.0048}          \\
Forestcover & 1.8694          & 3.1008          & 2.2987          & 1.1639 & 4.4611          & 2.1044     & \textbf{1.0021}          \\
Satimage-2 & \textbf{0.0223}           & 0.1333          & 0.0262          & 0.1127          & 0.1682          & 0.1459     & 0.0590 \\
Speech & 3.7235              & \textbf{3.3837} & 4.3909          & 3.8795          & 3.3979          & 3.4101     & 3.5571          \\
Pendigits & 0.5166            & 0.9059          & \textbf{0.4011} & 1.7072          & 0.7644          & 1.1826     & 1.5413          \\
Mammography & 0.8842          & 0.8469          & 1.0860          & \textbf{0.7733} & 2.5569          & 0.9512     & 1.5194          \\
Thyroid & \textbf{0.2041}             & 0.3360          & 0.6658          & 0.5627          & 0.6263          & 0.5105     & 0.4319 \\
Optdigits & \textbf{2.9540}            & 4.4296          & 6.6035          & 9.4286          & 7.6462          & 10.8933    & 11.1832  \\
Aloi & 0.4416                 & 0.6565          & 0.5413          & 0.3610  & 0.7000          & 0.5600     & \textbf{0.1796}     \\
Musk & \textbf{0.0000}                & 0.0624          & \textbf{0.0000}  & 0.0700         & 0.0324          & 0.0936     & 1.1902  \\
Vowels & \textbf{0.6185}               & 2.9484          & 1.1686          & 2.4464          & 1.8476          & 3.0582     & 1.9993 \\
Glass & 2.0255                & 1.9655          & 8.9189          & 2.5699          & \textbf{1.2163} & 2.1848     & 1.6546          \\
Letter & \textbf{1.5452}               & 3.0189          & 2.0284          & 2.7598          & 3.8942          & 2.6408     & 2.5764 \\
Shuttle & 0.0581              & 0.0586          & 0.0224          & 0.0995          & \textbf{0.0148} & 0.0763     & 0.5323          \\
Annthyroid & \textbf{0.6083}           & 2.1320          & 4.2616          & 1.7105          & 4.5212          & 2.5178     & 1.6343 \\
Wine & \textbf{0.9046}                 & 5.3481          & 0.9627          & 5.7927          & 2.8905          & 5.2280     & 20.7055 \\
Cardio & 1.8058               & 1.1376          & \textbf{0.5295} & 0.7788          & 1.9693          & 0.6126     & 0.7731          \\
Vertebral & 5.5261            & 5.9428          & 2.4988          & 5.7639          & \textbf{2.0460}  & 7.3348     & 3.1496          \\
Satellite & 0.6961            & 2.2929          & 1.0873          & 1.5878          & 1.2735          & 1.7396     & \textbf{0.6773} \\
Pima & \textbf{0.7969}                 & 1.4777          & 3.0418          & 1.8814          & 1.0238          & 1.8826     & 1.6276 \\
Breastw & 0.2262              & 0.1337          & 3.2579          & 0.2406          & \textbf{0.0668} & 0.1825     & 1.4448          \\
Ionosphere & 0.5082           & 0.5166          & 0.5804          & 0.9741          & \textbf{0.4167} & 0.8666     & 0.4793          \\ \midrule
Average    & \textbf{1.1206} & 1.7538          & 1.8808          & 1.9106          & 1.7697          & 2.0631     & 2.4418 \\
Rank        & \textbf{2.7083}          & 4.1250          & 3.9583          & 4.6250          & 4.0417     & 4.8333 & 3.6667  \\
Count        & \textbf{9}          & \textrm{1}          & \textrm{4}          & \textrm{2}          & \textrm{5}     & \textrm{0} & \textrm{4} \\ \bottomrule
\end{tabular}
\end{table}

\begin{figure}[!t]
\label{fig:robust}
  \centering
  \includegraphics[width=1.0\textwidth]{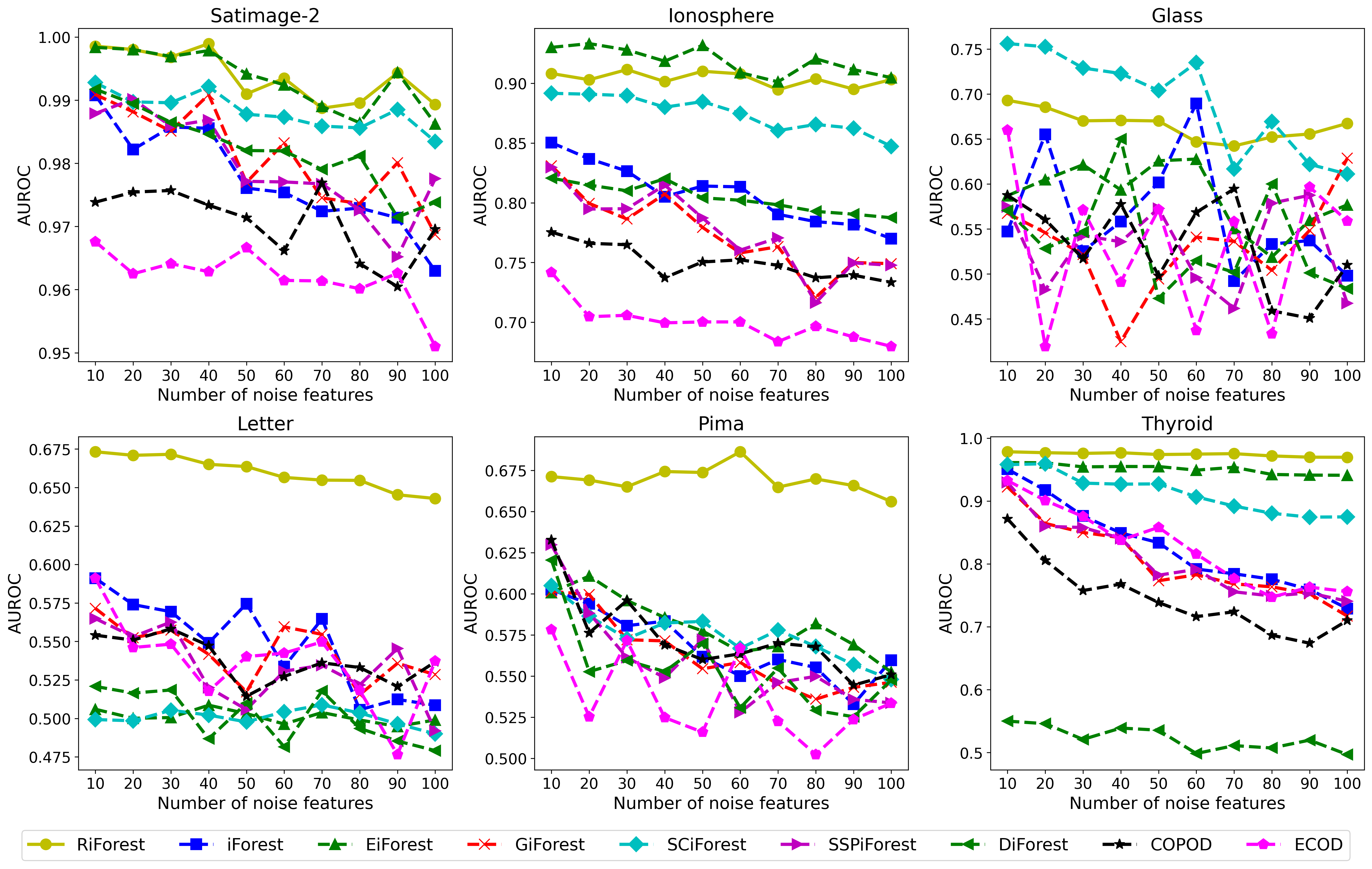}
  \caption{Results of the robustness experiments}\label{fig:robust}
\end{figure}

Finally, to statistically compare the outlier detection models in terms of anomaly detection performance and stability, we conducted a Friedman test \cite{friedman1940a} followed by Nemenyi's post-hoc test \cite{demsar2006statistical}. The Friedman test, a non-parametric test based on the rankings of algorithms across datasets, was performed using the AUROC and CV results from Table~\ref{tab:AUROC} and Table~\ref{tab:CV}. The test yielded p-values of 0.0083 for AUROC and 0.0239 for CV, allowing us to reject the null hypothesis of ``no statistical difference among algorithms'' at a significance level of 0.05. Subsequently, we proceeded with the post-hoc analysis.

For AUROC, RiForest performed statistically significantly better than all other algorithms except EiForest, iForest, and SCiForest. EiForest, which had the second-highest average rank after RiForest, showed no significant differences with any other iForest-based algorithm. COPOD and ECOD demonstrated statistically significantly lower anomaly detection performance compared to RiForest, EiForest, and iForest.

Regarding CV, RiForest showed statistically significantly better stability than all other iForest-based algorithms except DiForest, indicating that RiForest demonstrated superior stability regardless of dataset characteristics compared to the other iForest-based methods. DiForest, which had the second-highest average rank after RiForest, showed no significant differences with any of the other iForest-based algorithms.

In conclusion, RiForest statistically demonstrates superior anomaly detection performance and stability compared to other iForest-based algorithms.

\subsection{Robustness to noisy variables}
Next, Figure~\ref{fig:robust} presents the results of the robustness experiments involving noise. Among the six datasets used in these experiments, RiForest displayed the least performance variation when noise variables were added to the Ionosphere, Glass, Pima, and Thyroid datasets. Although SCiForest showed greater robustness than RiForest in the Satimage-2 and Letter datasets, RiForest also consistently maintained a very high performance level without severe degradation as the number of noise variables increased in these datasets. Among the existing algorithms, SCiForest and EiForest demonstrated relatively high robustness, while iForest, GiForest, SSPiForest, and DiForest---relying on entirely random hyperplane selection or random representations---exhibited significant performance degradation as the number of noise variables increased, making them highly susceptible to irrelevant variables. Similarly, COPOD and ECOD lacked robustness to noise variables because they incorporated the probability distributions of these irrelevant variables into their anomaly detection processes.

\begin{figure}[!tb]
\begin{minipage}{\textwidth}
\begin{center}
\includegraphics[width=0.7\linewidth]{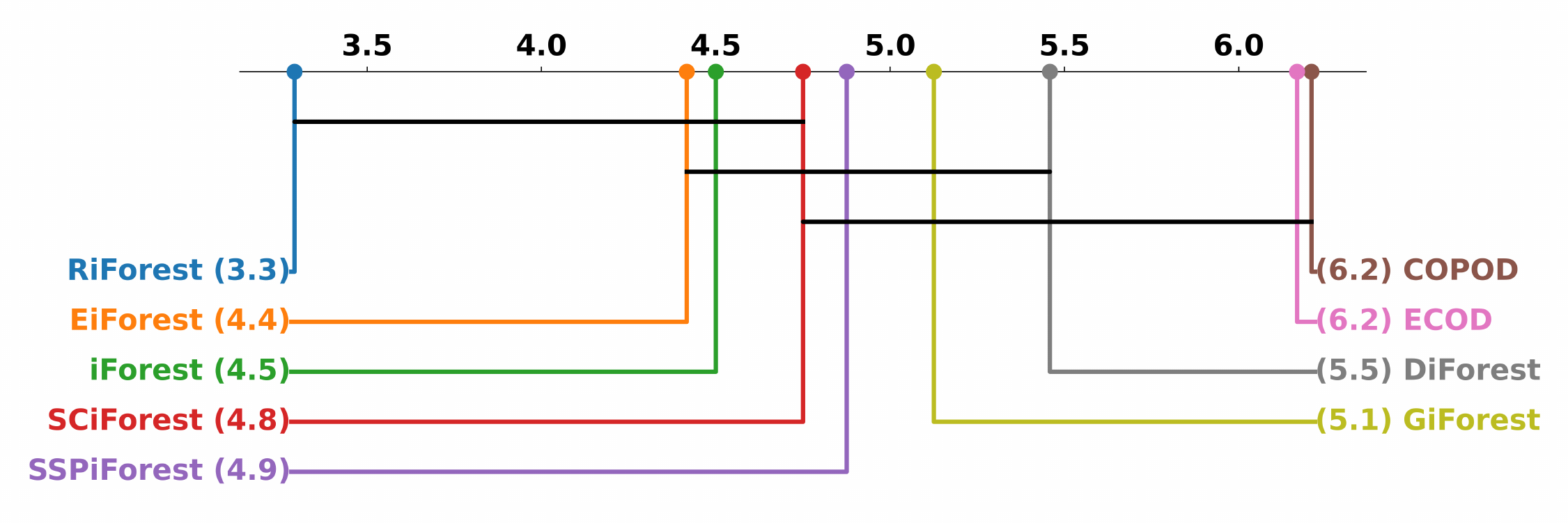} \\
    (a) AUROC
\end{center}
\end{minipage}
\begin{minipage}{\textwidth}
\begin{center}
\includegraphics[width=0.7\linewidth]{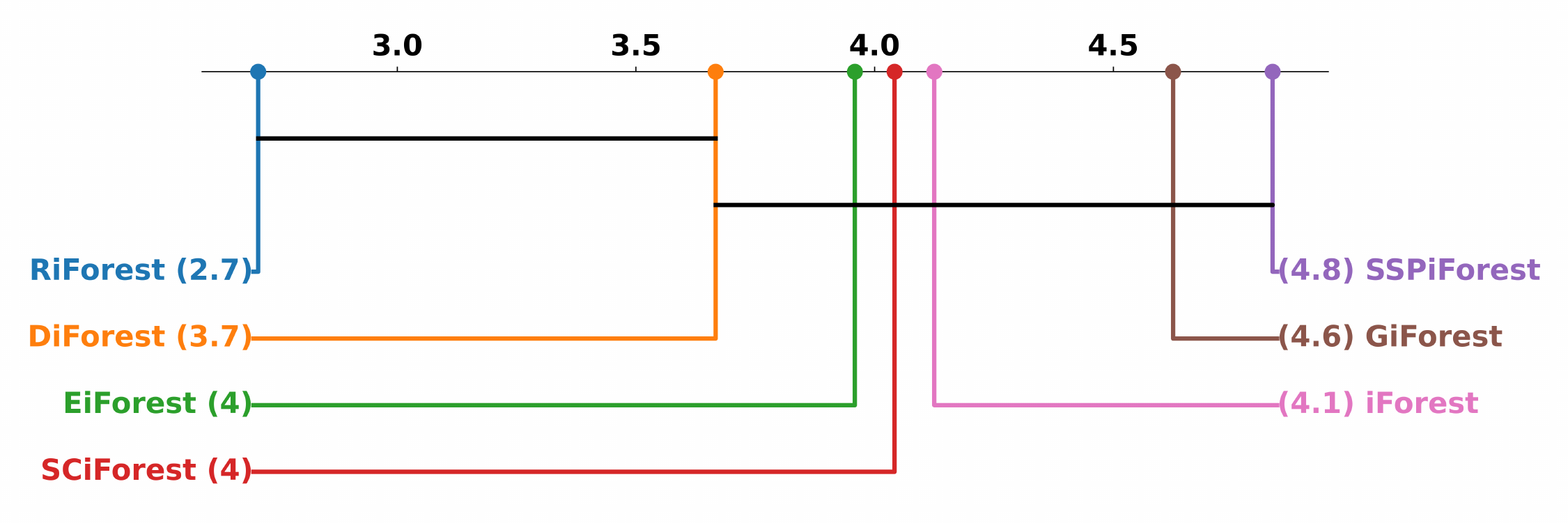} \\
    (b) CV
\end{center}
\end{minipage} 
  \caption{Nemenyi test figures on AUROC and CV}\label{fig:posthoc}
\end{figure}

\begin{figure}[!tb]

  \centering
  \includegraphics[width=0.9\textwidth]{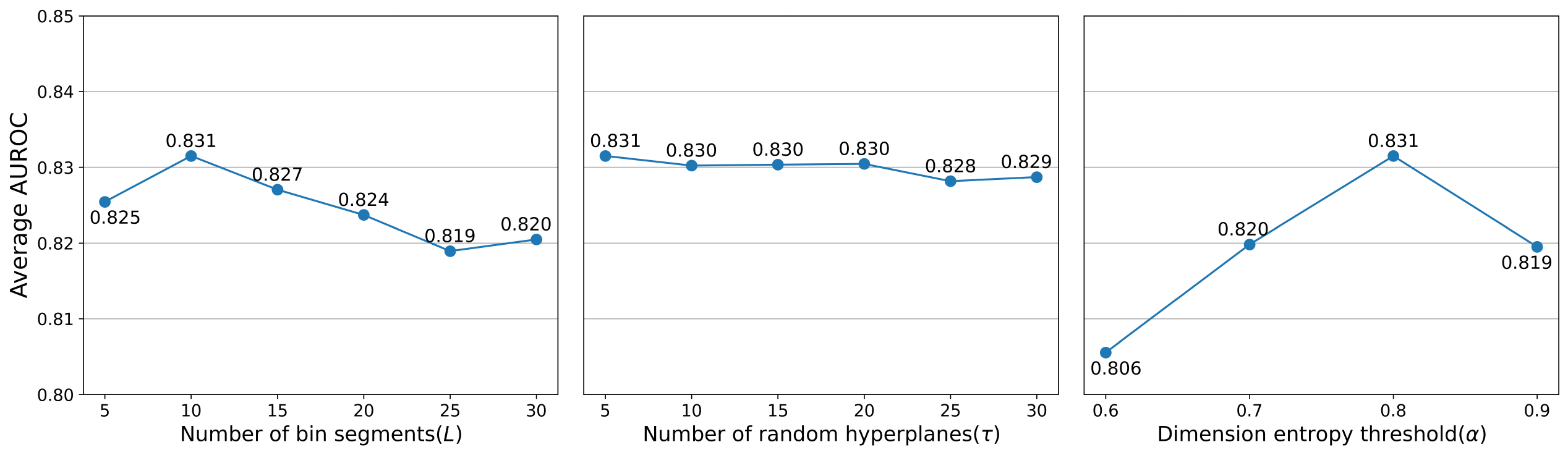}
  \caption{Hyperparameter analysis}\label{fig:hyper}
\end{figure}

\begin{table}[!tb]
\centering
\caption{Ablation study} \label{tab:ablation}
\footnotesize
\begin{tabular}{@{}lccccc@{}}
\toprule
            & w/o $pl$ & w/o RH & w/ RS & w/ BS & RiForest \\ \midrule
Avg AUROC        & 0.8308   & 0.8302   & 0.8237 & 0.8185 & 0.8315       \\
Avg $IR$ (\%)  & 3.7275  & 3.6591 & 3.0097 & 2.4863 & 3.7897       \\
Std of $IR$ (\%) & 7.3200 & 7.5250 & 4.9864 & 6.9694 & 7.3001      \\ \bottomrule
\end{tabular}
\end{table}

\subsection{Ablation study}
To validate the effectiveness of the key components adopted in RiForest to enhance the performance of iForest, an ablation study was conducted, and the results can be found in Table~\ref{tab:ablation}. The ablation models were compared using the mean AUROC, mean improvement rate, and standard deviation of improvement rate across 24 datasets. For a fair comparison of improvement rates, $\overline{AUROC}(D)$ in Eq.~\eqref{eq:IR} was calculated as the mean AUROC value from the 8 baseline models for dataset $D$, and the improvement rates of the different ablation models were calculated using this $\overline{AUROC}(D)$.

The model labeled ``w/o $pl$'' refers to an ablation model in which, instead of applying the newly proposed path length as specified in line 13 of Algorithm~\ref{alg:tree}, each node is assigned a path length of 1. The ``w/o RH'' model is an ablation in which random hyperplanes are excluded from the hyperplane set by omitting line 6 of Algorithm~\ref{alg:tree}. The models labeled ``w/ RS'' and ``w/ BS'' use random splits and the empty region-based split from EiForest, respectively, in place of the valley emphasis method for branching as implemented in line 10 of Algorithm~\ref{alg:tree}.

Firstly, all four ablation models exhibited lower average AUROC and mean improvement rate values compared to RiForest. This suggests that the path length assignment strategy, the candidate hyperplane set construction, and the valley emphasis-based split method introduced in RiForest all contribute to its superior overall performance. Notably, the performance decline in the ``w/ RS'' and ``w/ BS'' models was more pronounced than in the other ablation models, indicating that the valley emphasis-based split method is more effective than alternative split strategies and is the most significant factor contributing to RiForest's high performance. Another noteworthy observation is that the ``w/o RH'' model displayed the highest standard deviation in improvement rate, further confirming that incorporating random hyperplanes alongside existing attributes enables RiForest to consistently achieve high performance across diverse datasets without significant degradation, regardless of dataset variations.

\subsection{Hyperparameter analysis}\label{sec:hyper}
Figure~\ref{fig:hyper} illustrates the impact of RiForest’s hyperparameters on anomaly detection performance, measured by the average AUROC across 24 datasets. First, the number of bin segments achieved optimal performance at 10. When the number of bin segments was too low (5 or fewer) or too high (20 or more), performance decreased, likely due to the histogram's reduced ability to capture the overall distribution of the data. The number of random hyperplanes added to the candidate hyperplane set performed best at 5, with a general trend of declining performance as the number of random hyperplanes increased. However, performance variability due to the number of random hyperplanes was minimal. The dimension entropy threshold achieved optimal performance at 0.8 and exhibited the highest performance variability among all hyperparameters. When the dimension entropy threshold was below 0.7, irrelevant features were more likely to be included in the candidate set of split features, leading to performance degradation. Conversely, when the threshold was above 0.9, effective features for anomaly separation were less likely to be included in the candidate set of split features, also resulting in reduced performance.

\section{Conclusion}\label{sec:conclusion}
In this study, we introduced RiForest, an outlier detection algorithm based on the iForest framework, which demonstrates superior and more consistent performance compared to existing algorithms. RiForest incorporates two key strategies to achieve these results. First, it leverages both existing variables and random hyperplanes obtained through soft sparse random projection to select optimal hyperplanes for splitting. Second, it uses the valley emphasis method to determine ideal split points and adjust path lengths accordingly.

Through extensive experimentation on 24 benchmark datasets, RiForest demonstrated its effectiveness in improving anomaly detection performance, stability, and robustness to noise variables. It consistently achieved high AUROC values across diverse datasets compared to other iForest-based anomaly detection algorithms and the other types of anomaly detection algorithms such as COPOD and ECOD. Notably, when evaluating the improvement rate of different methods for each dataset, RiForest consistently showed positive improvement rates for most datasets, indicating its superior performance regardless of dataset characteristics. Additionally, RiForest maintained lower CV values and exhibited greater resilience to noisy features compared to other methods. Through an ablation study, the effectiveness of the components of RiForest was confirmed, and it was found that the valley emphasis method contributed most significantly to the performance improvement of RiForest.

Although the proposed RiForest represents a significant advancement over existing iForest-based algorithms, there remain areas for improvement. First, developing new criteria or rules for selecting effective features to better segregate anomalies is crucial, as feature selection significantly influences the performance of iForest-based algorithms. While this study employed the previously proposed dimension entropy, more effective feature selection methods could further enhance performance. Second, RiForest demonstrates relatively lower performance when most features exhibit a uniform distribution, indicating room for improvement in such scenarios. Specifically, when both original variables and features projected onto random hyperplanes are predominantly uniform, the candidate hyperplane set becomes limited, reducing splitting effectiveness. Addressing this limitation could pave the way for algorithms capable of achieving superior performance across a wider variety of datasets.

\section*{Declaration of Competing Interest}
The authors declare that they have no known competing financial interests or personal relationships that could have appeared to influence the work reported in this paper.

\section*{Acknowledgement}
This work was supported by the National Research Foundation of Korea(NRF) grant funded by the Korea government(MSIT) (No. RS-2023-00239374).

\bibliographystyle{unsrt}  
\bibliography{refs}

\end{document}